\documentclass[letterpaper, 10 pt, conference]{ieeeconf}  

\IEEEoverridecommandlockouts                              

\usepackage{graphics} 
\usepackage{epsfig} 
\usepackage{times} 
\usepackage{amsmath} 
\usepackage{amssymb}  
\usepackage{balance} 
\usepackage{algorithm}
\usepackage{amsfonts}
\let\labelindent\relax
\usepackage[inline]{enumitem}
\usepackage[noend]{algpseudocode}
\usepackage{color,subfigure}
\usepackage{booktabs}
\usepackage{color}
\usepackage{blindtext}
\usepackage[pagebackref=false,breaklinks=true,colorlinks=true,bookmarks=false,linkcolor={black},urlcolor={black},citecolor={green}]{hyperref} 

\usepackage{url}
\usepackage{pifont}
\usepackage{booktabs} 
\usepackage{etoolbox}
\makeatletter
\patchcmd{\@makecaption}
  {\scshape}
  {}
  {}
  {}
\makeatletter
\patchcmd{\@makecaption}
  {\\}
  {.\ }
  {}
  {}
\makeatother

\title{\LARGE \bf
Boosting Feedback Efficiency of Interactive Reinforcement Learning  by Adaptive Learning from Scores
}

\author{Shukai Liu$^{1}$, Chenming Wu$^{2\dagger}$, Ying Li$^{3}$ and Liangjun Zhang$^2$
\thanks{$\dagger$ Corresponding author}
\thanks{$^{1}$S. Liu is with the Department of Engineering Mathematics, University of Bristol, UK.
        {\tt\small lshukai22@gmail.com}}%
\thanks{$^{2}$C. Wu and L. Zhang are with RAL, Baidu Research.
        {\tt\small \{wuchenming, liangjunzhang\}@baidu.com}}%
\thanks{$^{3}$Y. Li is with the School of Mechanical Engineering, Beijing Institute of Technology, Beijing, China.
        {\tt\small ying.li@bit.edu.cn}}
}

\begin{document}

\maketitle
\thispagestyle{empty}
\pagestyle{empty}

\begin{abstract}
Interactive reinforcement learning has shown promise in learning complex robotic tasks. However, the process can be human-intensive due to the requirement of a large amount of interactive feedback. This paper presents a new method that uses scores provided by humans instead of pairwise preferences to improve the feedback efficiency of interactive reinforcement learning. Our key insight is that scores can yield significantly more data than pairwise preferences. Specifically, we require a teacher to interactively score the full trajectories of an agent to train a behavioral policy in a sparse reward environment. To avoid unstable scores given by humans negatively impacting the training process, we propose an adaptive learning scheme. This enables the learning paradigm to be insensitive to imperfect or unreliable scores. We extensively evaluate our method for robotic locomotion and manipulation tasks. The results show that the proposed method can efficiently learn near-optimal policies by adaptive learning from scores while requiring less feedback compared to pairwise preference learning methods. The source codes are publicly available at \href{https://github.com/SSKKai/Interactive-Scoring-IRL}{https://github.com/SSKKai/Interactive-Scoring-IRL}.

\end{abstract}

\section{Introduction} 
\label{sec:intro}


Deep Reinforcement Learning (DRL) has made remarkable progress in addressing robotic control tasks, such as legged robot locomotion \cite{3yang2020multi} and robotic manipulation \cite{6liu2021deep}. However, formulating an accurate reward function for a specific task can pose a significant challenge. Sparse reward functions are frequently employed for their simplicity, but their absence of reward signals can result in longer exploration and training time \cite{7vecerik2017leveraging} and lower success rates \cite{8mohtasib2021study}. In general, tasks with high complexity and high-dimensional continuous state-action spaces benefit from denser reward signals. Nonetheless, creating a reward function for autonomous agents necessitates domain expertise, which can be challenging for non-experts. Furthermore, hand-crafted rewards can be vulnerable to local optima or unexpected ways of achieving high numerical returns \cite{9hadfield2017inverse}.


\begin{figure}[!t]
\centering
\includegraphics[width=3.2in]{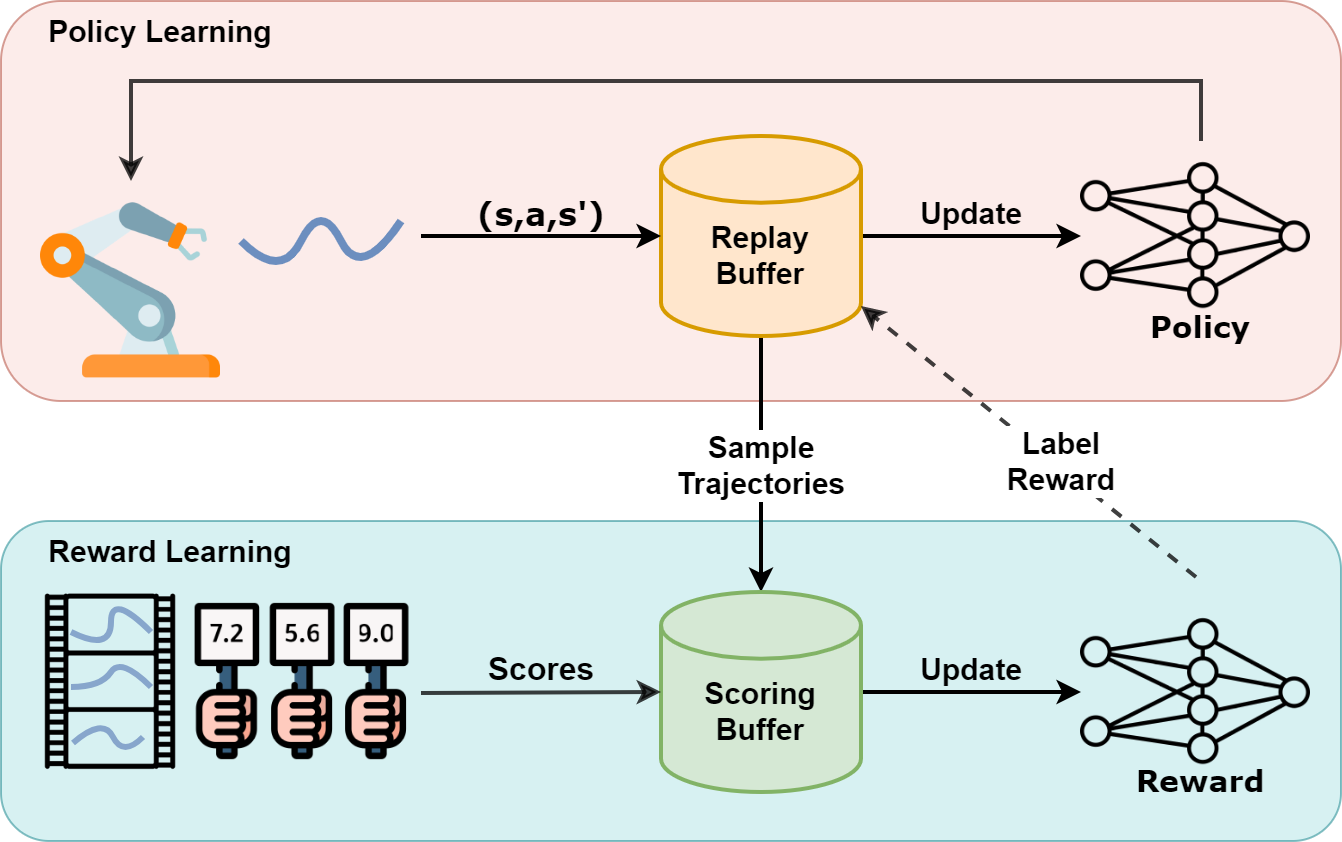}
\caption{Our proposed approach consists of two main components: policy learning and reward learning. The policy learning component updates itself through an off-policy RL algorithm to maximize the sum of predicted rewards generated by a reward network. Meanwhile, the reward learning component periodically samples trajectories, asks teachers to score them and optimizes its network.}
\vspace{-10pt}
\label{figure1}
\end{figure}



Inverse reinforcement learning (IRL) is a problem in which the reward function is inferred from expert demonstration trajectories \cite{arora2021survey}. This eliminates the need for tedious reward engineering and makes it possible to learn reward functions from expert demonstrations \cite{10ng2000algorithms, 11abbeel2004apprenticeship}. However, IRL typically requires optimal demonstrations, while those demonstrated by people are often sub-optimal.
To address this issue, a new method called Trajectory-ranked Reward Extrapolation (T-REX) was recently proposed. T-REX seeks to improve demonstrations by learning rewards from sequences of suboptimal ranked demonstrations. It attempts to convert those rankings into a set of pairwise preferences to train the policy network~\cite{brown2019extrapolating}. However, T-REX still requires a large number of demonstrations to train the policy, which can be challenging in tasks where providing demonstrations at such a scale is difficult, even though optimality is not required.

In a recent study, PEBBLE~\cite{lee2021pebble} proposed an off-policy interactive RL algorithm to train a reward network and a policy network simultaneously from queried pairwise preferences. Teachers provide real-time pairwise feedback to supervise the learning process in the most efficient direction. However, this approach presents three major issues when providing feedback by assigning one-hot preference labels to two trajectories:
1) Two trajectories can be compared only when they have been paired together, making it difficult to gain a broader understanding of the relationship between individual and overall sampled trajectories.
2) Forcing the teacher to prioritize a better trajectory can sometimes become a burden and harm human-in-the-loop training.
3) To increase the number of training examples, partial trajectories (i.e., partitioning a full trajectory into segments) are used rather than full trajectories, which can make evaluating pairwise preferences more ambiguous for the teacher.

Our aim is to enhance feedback efficiency in interactive reinforcement learning. To achieve this, we suggest utilizing scores instead of pairwise preferences as the signals for interacting with RL agents. Moreover, we put forward an adaptive learning scheme to make the training process smoother and more stable.
This includes adaptive network optimization to smoothly update network parameters from score data and adaptive trajectory sampling to mine useful trajectories for teachers to evaluate, making our methodology less sensitive to imperfect or unreliable teacher inputs. By interchangeably giving feedback and adjusting the reward function, we continuously optimize both the reward and policy networks. Furthermore, we implement a scoring graphical user interface (GUI), showing the most relevant trajectories from the previously scored ones when a new trajectory is scored to support users in providing consistent scores. Teachers are allowed to amend and correct previous scores during the training process.
An overview of our proposed framework is illustrated in Fig.~\ref{figure1}. The main contributions of this paper are summarized as follows.
\begin{enumerate}
\item We develop an interactive RL method that enables the agent to learn both policy and reward simultaneously using a score-based approach. The RL agent proactively requests scores from teachers for complete trajectories, which results in requiring less feedback compared to pairwise-based methods.
\item We propose a method to tackle the problem of inaccuracies in scoring by introducing an adaptable learning approach that can withstand errors. Our proposed method also facilitates efficient learning of personalized and desired behaviors in situations where rewards are limited, based on the teachers' choices.
\end{enumerate}


\section{Related Work} 


\subsection{Inverse Reinforcement Learning}
IRL allows the agent to better understand tasks and the environment, and learn an optimal policy using the reward via RL methods \cite{sutton1998introduction}. However, classic IRL frameworks \cite{ng2000algorithms,11abbeel2004apprenticeship} assume that demonstrations are optimal and easy to obtain. Maximum entropy IRL \cite{ziebart2008maximum,wulfmeier2015maximum,finn2016guided} and Bayesian IRL \cite{ramachandran2007bayesian,okal2016learning} are more robust to limited and stochastic suboptimality, but they cannot produce a policy better than the demonstration, thus their performance still highly relies on the quality of the demonstration. In \cite{coates2008learning}, a generative model is learned from a large number of suboptimal demonstrations to produce noise-free trajectories. In \cite{shiarlis2016inverse}, the reward function is formed as a linear combination of known features, and suboptimal demonstrations are utilized by learning rewards from trajectories labeled as success or failure. The method proposed in \cite{choi2019robust} is robust to a limited number of non-optimal demonstrations but still requires many expert demonstrations to discriminate the suboptimal ones. However, these methods can be challenging to apply in real-world scenarios where demonstrations are scarce, expensive, or suboptimal.

\subsection{Learning from Evaluative Feedback}
Evaluative feedback is a value given by a human teacher that rates the quality of the agent's behavior, which is easier for humans to provide compared to demonstrations. The TAMER framework \cite{knox2009interactively} interprets evaluative feedback as a $Q^*(s,a)$ function of RL, and makes the agent act greedily according to it. Meanwhile, the COACH framework \cite{macglashan2017interactive} interprets human feedback as the advantage function $A^{\pi}(s,a)$ of policy gradient update. In the policy shaping framework \cite{griffith2013policy,cederborg2015policy}, evaluative feedback is considered an optimality label of the action. 
Providing evaluative feedback for entire trajectories can give each trajectory a global evaluation of its quality and, therefore, more accessible generalization. \cite{el2016score,burchfiel2016distance} annotate each trajectory with a numeric score of a human teacher's global performance and leverage the IRL framework to learn a reward by minimizing the distance between human-provided and predicted scores. Although these methods show robustness to scoring errors, they cannot deal with suboptimal and high-dimensional IRL tasks.

Evaluative feedback is useful for handling non-optimality. CEILing \cite{chisari2022correct} labels all state-action pairs with binary feedback evaluative feedback, then directly learns a Gaussian distributed policy by reinforcing the good ones while ignoring the bad ones. Similarly, \cite{mourad2020learning} proposes IRLDC, which uses binary evaluative feedback to label each state-action pair. These methods still require a few demonstrations or correct feedback.

\subsection{Preference-based Reinforcement Learning}
\cite{christiano2017deep} introduces the preference-based DRL framework, which can learn from pairwise preferences over the agent's current behaviors that the human teacher actively provides during training. This approach is on-policy that needs the human teacher to constantly provide preference feedback during the training process. Thus, \cite{brown2019extrapolating}. extends this framework and proposes T-REX, which learns the reward function from the pairwise preferences derived from a set of pre-collected ranked demonstrations, then applies the reward to RL for policy learning. T-REX allows the learned reward to extrapolate beyond the demonstrations and achieve better-than-demonstrator performance. D-REX \cite{brown2020better} and SSRR \cite{chen2020learning} extend the learning-from-ranking framework by automatically generating ranked trajectories via noise injection. However, the need for demonstrations still exists.

Myers et al. \cite{myers2022learning} proposed a robot learning method that can learn multimodal rewards from multiple active ranking queries by multiple experts. PEBBLE \cite{lee2021pebble} presented an interactive preference learning method that enables users to give preference feedback directly on the behavior of the RL agent, thus eliminating the need for demonstrations. PEBBLE introduces the off-policy learning framework to reuse data and follows the feedback form of pairwise preferences between partial trajectories for the sample efficiency as in \cite{christiano2017deep}. To improve feedback efficiency, \cite{lee2021b} investigates the query selection and policy initialization. \cite{liang2022reward} presents an exploration method to collect more diverse experiences. \cite{wang2022feedback} introduces this learning scheme for socially aware robot navigation and reduces the amount of preference feedback from humans by collecting expert demonstrations. \cite{park2022surf} further increases the feedback efficiency by inferring pseudo-labels on a large number of unlabeled samples with data augmentation, while we annotate global scores to the agent's past experience. In our work, we annotate global scores to the agent's past experiences and demonstrate that this scoring feedback scheme can substantially reduce the amount of required feedback and better fit the off-policy framework.
 
 


\section{Methodology}

Our proposed framework can be broken down into two processes. First, the RL agent interacts with the environment to create new trajectories {to be scored}. Second, an off-policy DRL algorithm is applied to update the agent's policy $\pi_\psi$ in order to maximize the expectation of the predicted reward generated by $\hat{r}_\theta$. Additionally, the teacher reviews the sampled trajectories through video replay and scores them at a frequency of $f$ during the RL training process. The scored trajectories $(\tau, s)$ are stored in the scoring buffer $\mathcal{D}$ to update the reward network. The agent then deduces the teacher's preference from the score difference and updates the reward network accordingly. The updated reward network guides the agent to generate better trajectories. Scoring these trajectories can lead to a more comprehensive reward network, and the agent learns the policy and reward simultaneously.

\subsection{Adaptive Learning from Scores}

The reward function is trained as a neural network, which we refer to as $\hat{r}_\theta$. Users can choose either states or state-action pairs as input. Our approach utilizes two replay buffers: one for the RL part to store the state-action transitions, and the other for reward learning to store the trajectories and their scores. For policy learning, the only difference between our method and vanilla RL algorithms is that the rewards are produced by the reward network. As a result, our approach can be applied to most off-policy RL algorithms while maintaining their core functionality.

Note that the reward function is dynamically updated during training, which can cause inconsistency in off-policy RL since previously generated rewards may not match the latest reward functions. To address this issue, we adopt the approach proposed in PEBBLE~\cite{lee2021pebble} and relabel the replay buffer each time we update the reward. Storing all scored trajectories in the scoring buffer allows for off-policy learning in reward learning. This allows newly scored trajectories to be compared with previously scored trajectories, which significantly improves the utilization rate of human feedback. Moreover, the scoring buffer allows the teacher to access previous scores during training and correct them if they change their minds, making reward learning more robust.

\begin{figure*}[!ht]
	\centering
    \subfigure[Mujoco Tasks]{%
    \includegraphics[height=4.2cm]{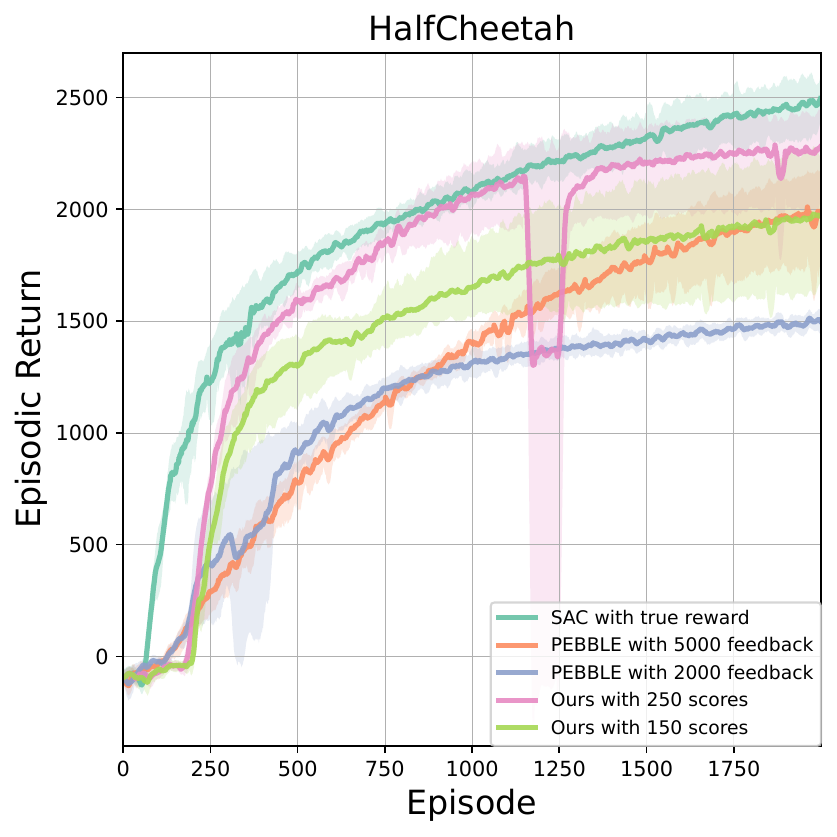}
    \includegraphics[height=4.2cm]{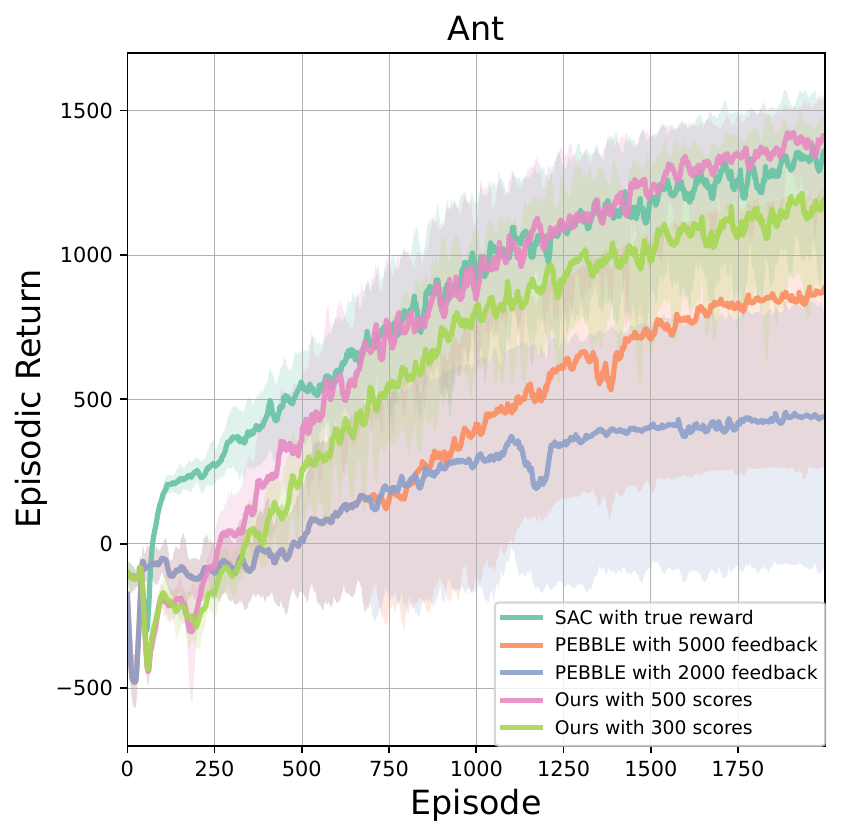}\label{2a}
    }
    \subfigure[Metaworld Tasks]{%
    \includegraphics[height=4.2cm]{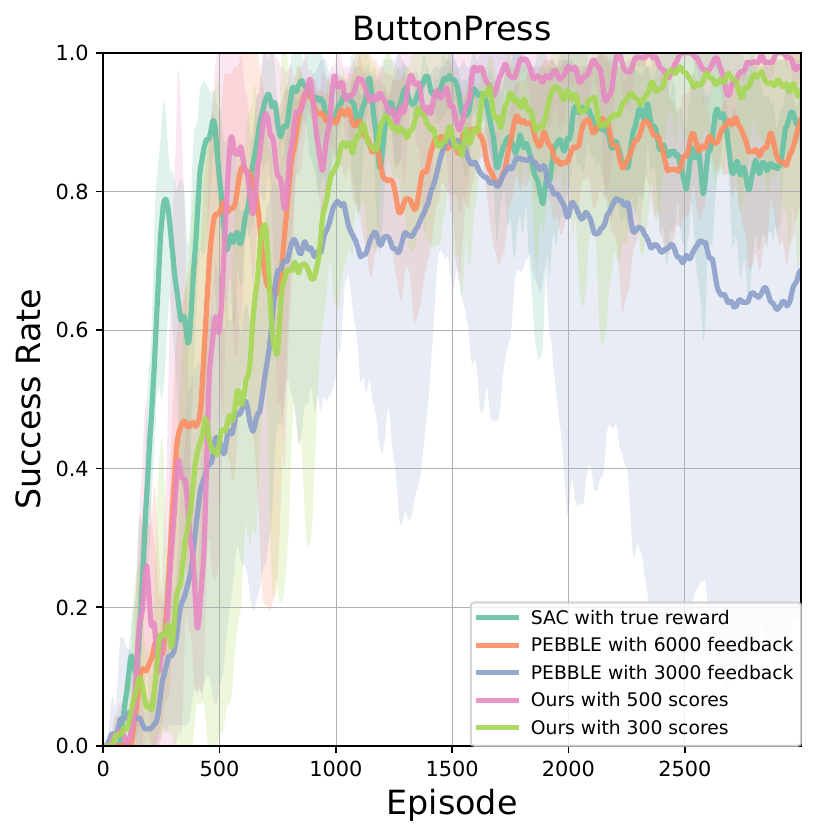}
    \includegraphics[height=4.2cm]{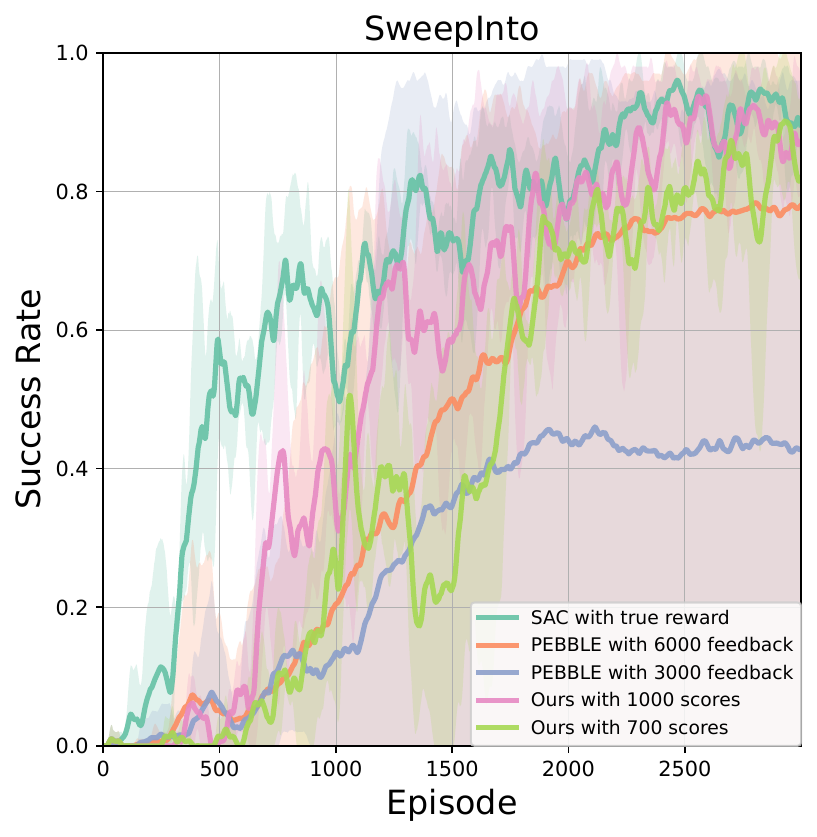}\label{2b}
    }
    \caption{The graphs display the learning progress for SAC, PEBBLE, and our method in two different simulations: HalfCheetah and Ant for movement tasks, and ButtonPress and SweepInto for robotics tasks. The data is measured by the success rate for the Metaworld environment and ground true return for Mujoco. The solid lines indicate the mean, while the shaded regions show the minimum and maximum range across three runs for all figures.}
    \label{figure2} 
    \vspace{-16pt}
\end{figure*}

\vspace{4pt} \noindent \textbf{Adaptive Network Optimizing.} Given a set of trajectories $\tau_1, \tau_2,\ldots, \tau_m$ and their corresponding scores $s_1, s_2,\ldots,s_m$, our goal is to parameterize a reward function $\hat{r}_\theta$ to infer the underlying reward from scores and output the reward value that matches user's evaluation standard, such that $\sum_{\tau_i}\hat{r}_\theta < \sum_{\tau_j}\hat{r}_\theta$ when $s_i < s_j$. This problem can be regarded as a learning-to-rank problem~\cite{cao2007learning}, to optimize the following equation:

\begin{equation}
   \min {\rm{U}}({\rm \bf sorted}(\hat{r}_\theta(\tau_1), ..., \hat{r}_\theta(\tau_{|\mathcal{D}|})), \textbf{y})
\end{equation}

\noindent where $\textbf{y}$ is the ground-truth index list and $\rm{U}$ is a binary function to evaluate if the ranked position is equivalent to the ground-truth position $y_i$ in $\textbf{y}$. In our work, we decide to solve this problem using the idea of pairwise learning because our major goal is to train a reward function, instead of a ranker that generates a descent permutation.
The user's preference over any pair of two trajectories is described by a distribution $\mu$, which can be derived by comparing their scores, e.g., $\mu=1$ if $s_i < s_j$. By following the Bradley-Terry and Luce-Shephard models of preferences~\cite{bradley1952rank,luce2012individual}, a preference predictor using the reward function $\hat{r}_\theta$ can be modeled as a softmax-normalized distribution as follows.

\begin{equation}\label{eq1}
    P(\tau_i \prec \tau_j) = \frac{e^{\sum_{\tau_j}\hat{r}_\theta}}{e^{\sum_{\tau_i}\hat{r}_\theta} + e^{\sum_{\tau_j}\hat{r}_\theta}}
\end{equation}

\noindent where $\tau_i \prec \tau_j$ denotes the trajectory $\tau_j$ is more preferred to the trajectory $\tau_j$. This equation demonstrates that the probability of preferring one trajectory to another is exponentially related to the predicted return of each trajectory. Thus, the parameterized reward function $\hat{r}_\theta$ can be learned via minimize the cross entropy loss between the predicted preference and the user's true preference as follows.

\begin{equation}\label{eq2}
    \mathcal{L} = -\sum_{(\tau_0, \tau_1, \mu) \in \mathcal{D}}[ \mu \log P(\tau_i \prec \tau_j) + (1-\mu) \log P(\tau_j \prec \tau_i) ]
\end{equation}

The preference distribution $\mu$ is usually a one-hot encoded label. Although this can learn reward effectively on correct labels, it may suffer from poor performance when there are wrong labels in the database $\mathcal{D}$. Unfortunately, it is nearly impossible for a human user to score a large number of trajectories perfectly. To strengthen the robustness against scoring error, we use the label smoothing method \cite{szegedy2016rethinking} to convert the hard label $\mu$ to soft label $\tilde{\mu}$ using $\tilde{\mu} = (1-\alpha)\mu + \alpha/K$, where $\alpha \in [0,1]$ is a constant factor that indicates the smoothing strength for one-hot label and $K$ denotes the number of labels, in our case, $K=2$. However, in our setting, using a constant smoothing strength for all pairwise labels may not be ideal because it ignores the relative relationship implicit in the score differences. It is intuitive that for a trajectory pair, the larger the scores differ, the more confident that one trajectory is better than the other. Thus, to better exploit the information from human scores, we make the smoothing strength adaptive to the score differences by $\alpha = 1/(|s_i-s_j|+\lambda)^{2}$ where {$\lambda>1$} is a hyperparameter, and we set it as $2$ in all our experiments. The adaptive $\alpha$ makes the label $\tilde{\mu}$ closer to 0 or 1 when pairwise trajectories significantly differ in the score and approach 0.5 when the scores are similar. Hence, the soft label $\tilde{\mu}$ is computed as

\begin{equation}\label{eq3}
    \tilde{\mu} = (1-\frac{1}{(|s_i-s_j|+\lambda)^{2}})\mu + \frac{1}{K(|s_i-s_j|+\lambda)^{2}}
\end{equation}

\noindent \textbf{Adaptive Trajectory Sampling.} The process of learning through rewards includes two sampling procedures. The first one involves gathering newly created trajectories from the RL agent, which are then evaluated and given scores by teachers. The second procedure involves selecting a batch of trajectory pairs from the scoring buffer, which stores all scored trajectories.

To improve the training of our RL agent, we ask the user to rate the newly generated trajectories. These ratings are stored in the scoring buffer called $\mathcal{D}$ as scored trajectories $(\tau, {s})$. However, asking for scores for all trajectories can be overwhelming. Therefore, we aim to choose the most informative scoring queries. This way, even if only a few newly generated trajectories are scored each time, they are enough to train the appropriate reward. We employ the $k$-means clustering algorithm to automatically select trajectories with high variance in performance, which are then approximated by the predicted rewards. To select $k$ trajectories from a set of newly generated ones for evaluation, we use the reward network to compute the episodic return of each trajectory. Next, we run $k$-means clustering on these returns and choose the $k$ trajectories whose returns are closest to each $k$ centroid.

For $n$ scored trajectories $(\tau, s)$ in the scoring buffer $\mathcal{D}$, we notice that not all {the trajectory pairs} of them can lead to effective reward update. To address this, we explore methods for sampling from the scoring buffer $\mathcal{D}$. An off-the-shelf sampling method is the entropy-based sampling adopted in PEBBLE~\cite{lee2021pebble}, which randomly samples a large batch of trajectories and seeks to maximize the entropy. However, we notice that when a trajectory with a higher score is sampled into $\mathcal{D}$, it should be compared more broadly with other trajectories. This allows the reward equation to learn what behaviors lead to higher scores. Unfortunately, entropy-based sampling cannot provide this capability. Inspired by the prioritized experience replay (PER) methods \cite{schaul2015prioritized},  we propose an alternative sampling methodology: either randomly selecting one trajectory in a pair or choosing based on a probability that increases with its score. The probability of each scored trajectory is computed according to its score as 
$P(i) = \frac{s_i^\beta}{\sum_n{s_n^\beta}}$, where $\beta$ is a hyperparameter that determines how much prioritization is assigned to a high-scored trajectory. And $n$ is the total number of trajectories in the scoring buffer. The comparison between our sampling method and entropy-based sampling is shown in the experimental section.

\section{Experiments}
\label{sec:Experiments}

\begin{figure} [t]
    \centering
    \subfigure[Effects of reward update methods]{%
    \includegraphics[height=4.2cm]{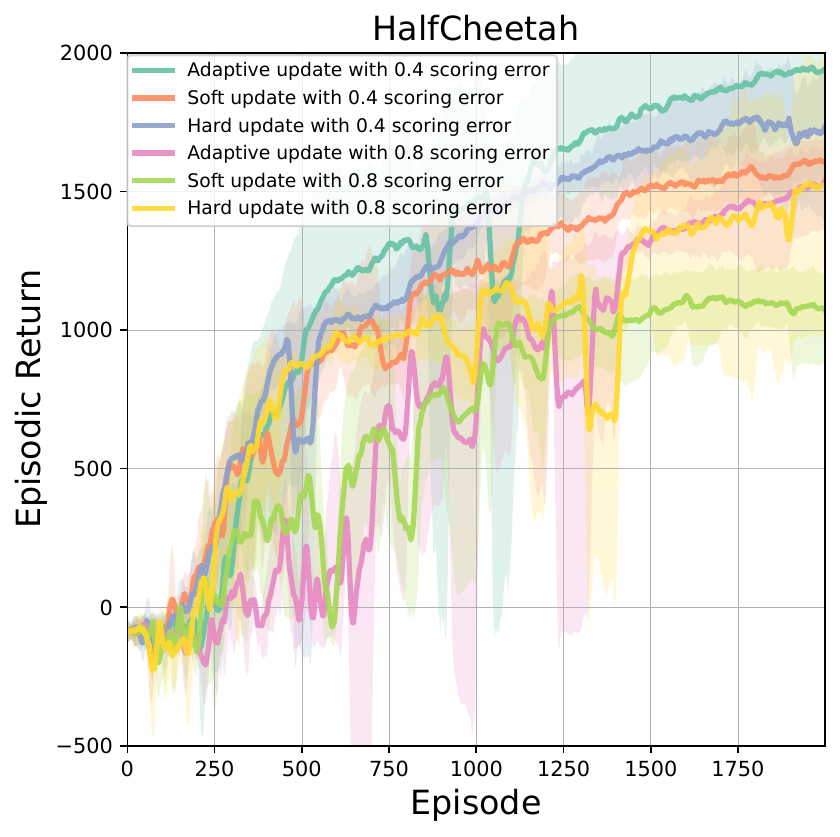}
    \includegraphics[height=4.2cm]{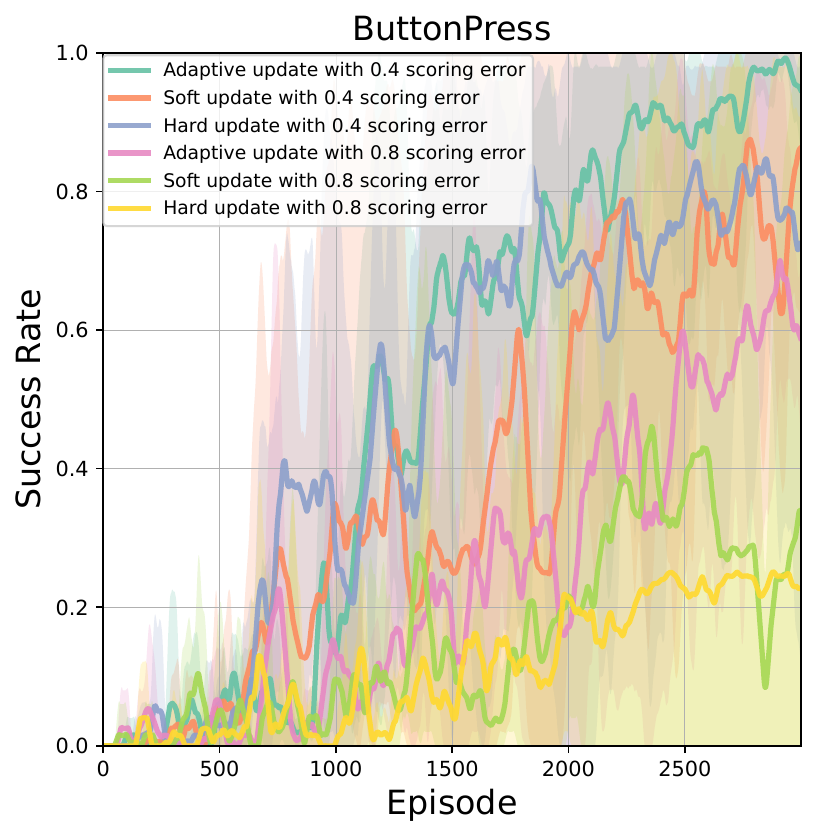}\label{3a}
    }
    \\
    \subfigure[Effects of sampling schemes]{%
    \includegraphics[height=4.2cm]{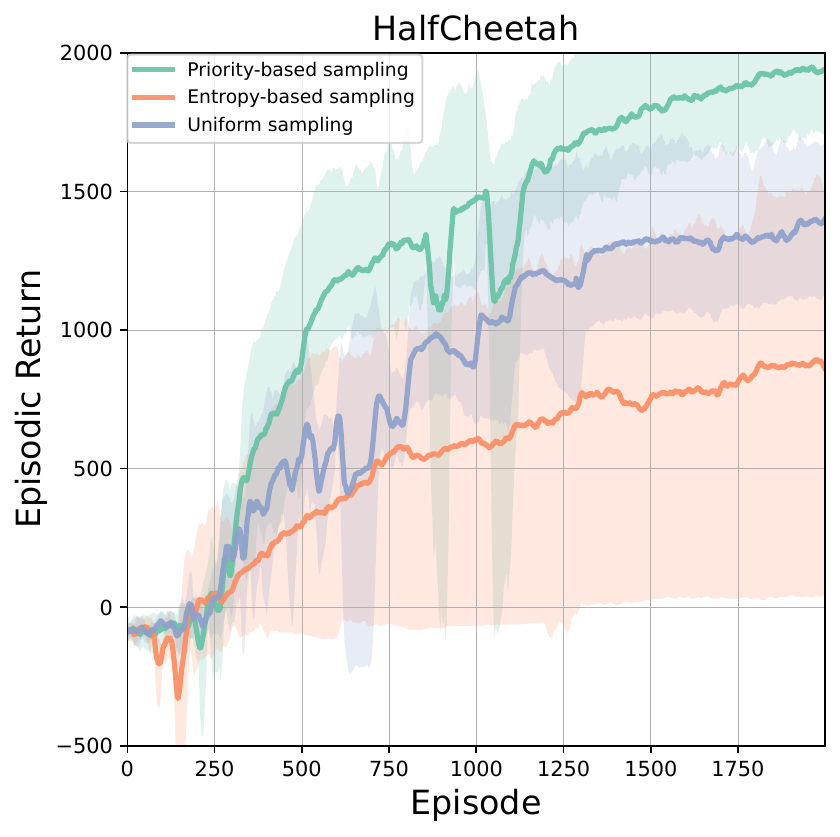}
    \includegraphics[height=4.2cm]{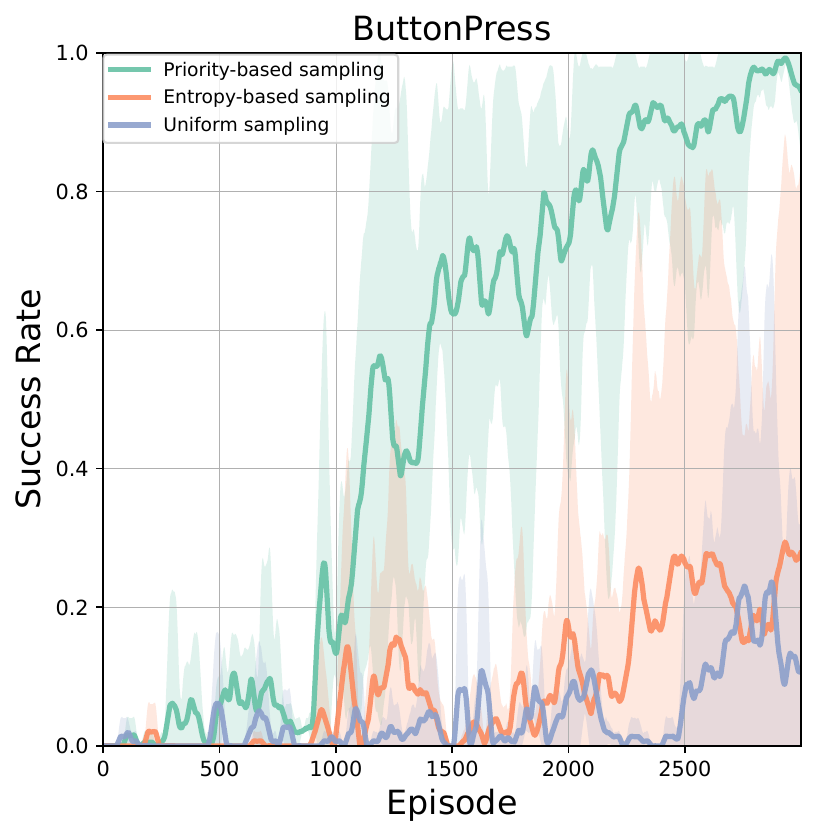}\label{3b}
    }
    \caption{Ablation study on HalfCheetah and ButtonPress. (a) The performance comparisons of different update reward methods under the scoring noise of 0.4 and 0.8. (b) Effects of sampling schemes to select training batch for reward update under the scoring noise of 0.4. We set a maximum of 400 scores for HalfCheetah and 1000 scores for ButtonPress.}
    \label{figure3} 
    \vspace{-10pt}
\end{figure}

\begin{figure} [t]
    \centering
    \subfigure[Analysis of episodic returns]{%
    \includegraphics[height=4.2cm]{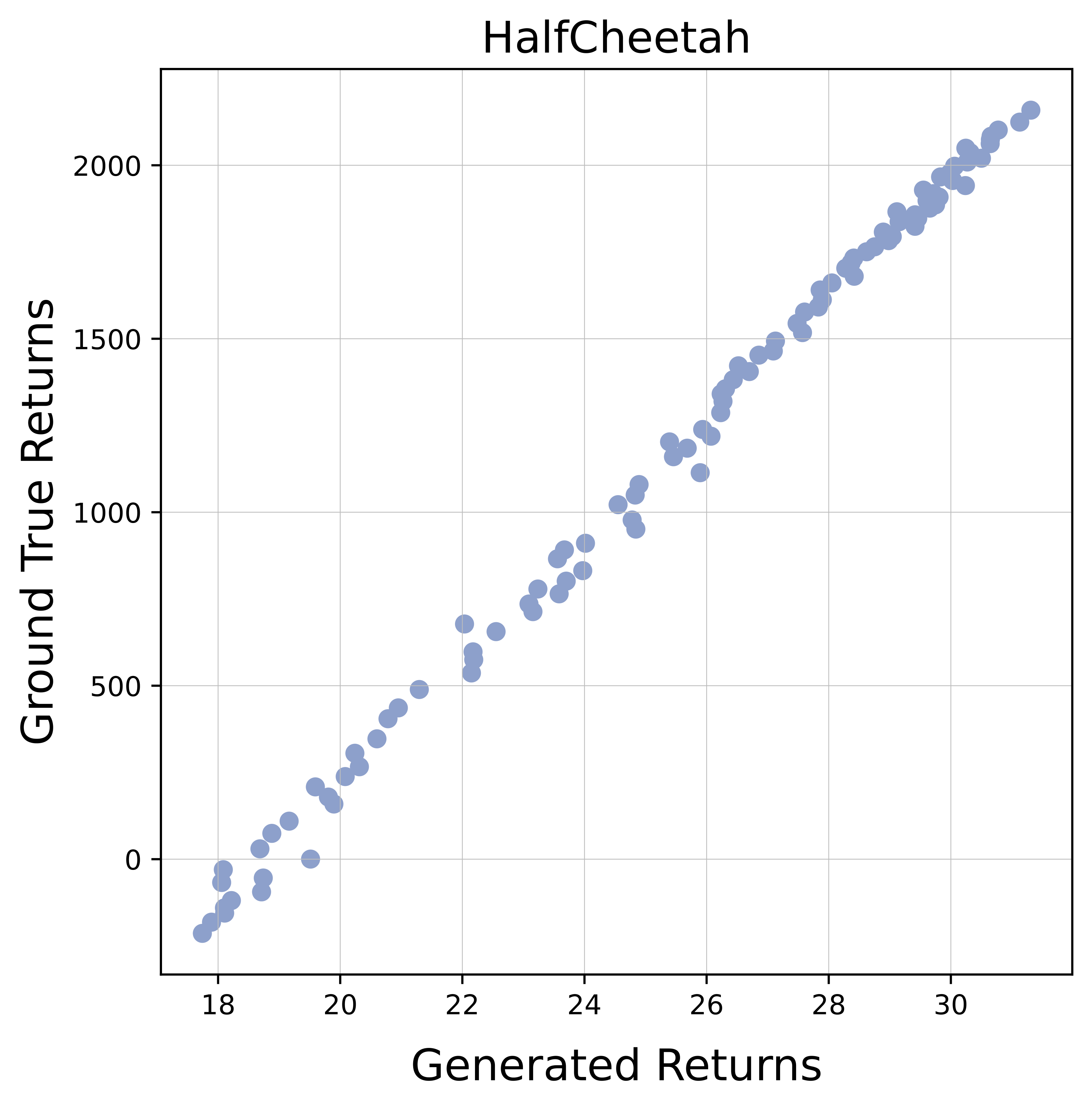}
    \includegraphics[height=4.2cm]{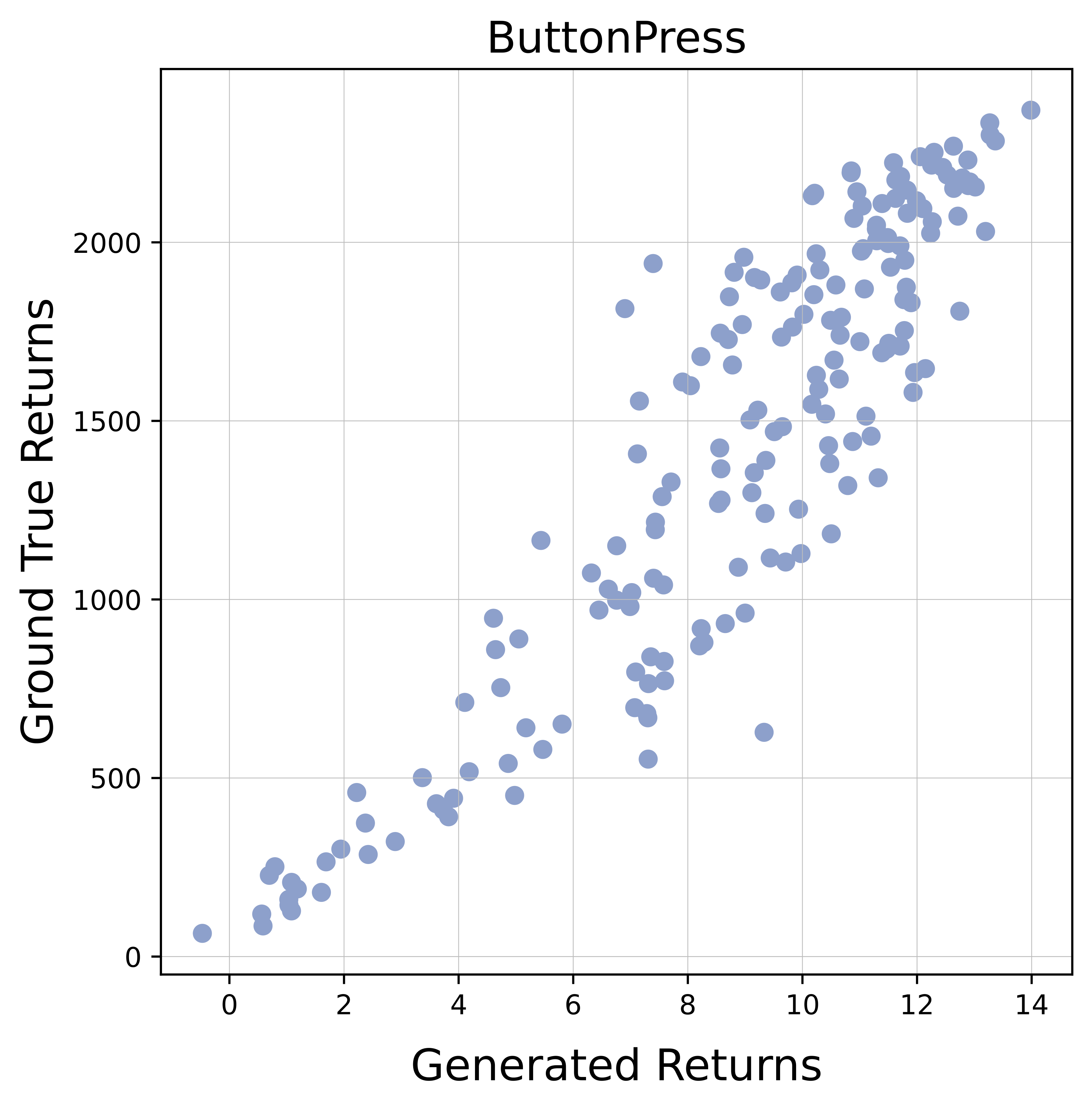}\label{4a}
    }
    \\
    \subfigure[Analysis of reward signals within a episode]{%
    \includegraphics[height=4.2cm]{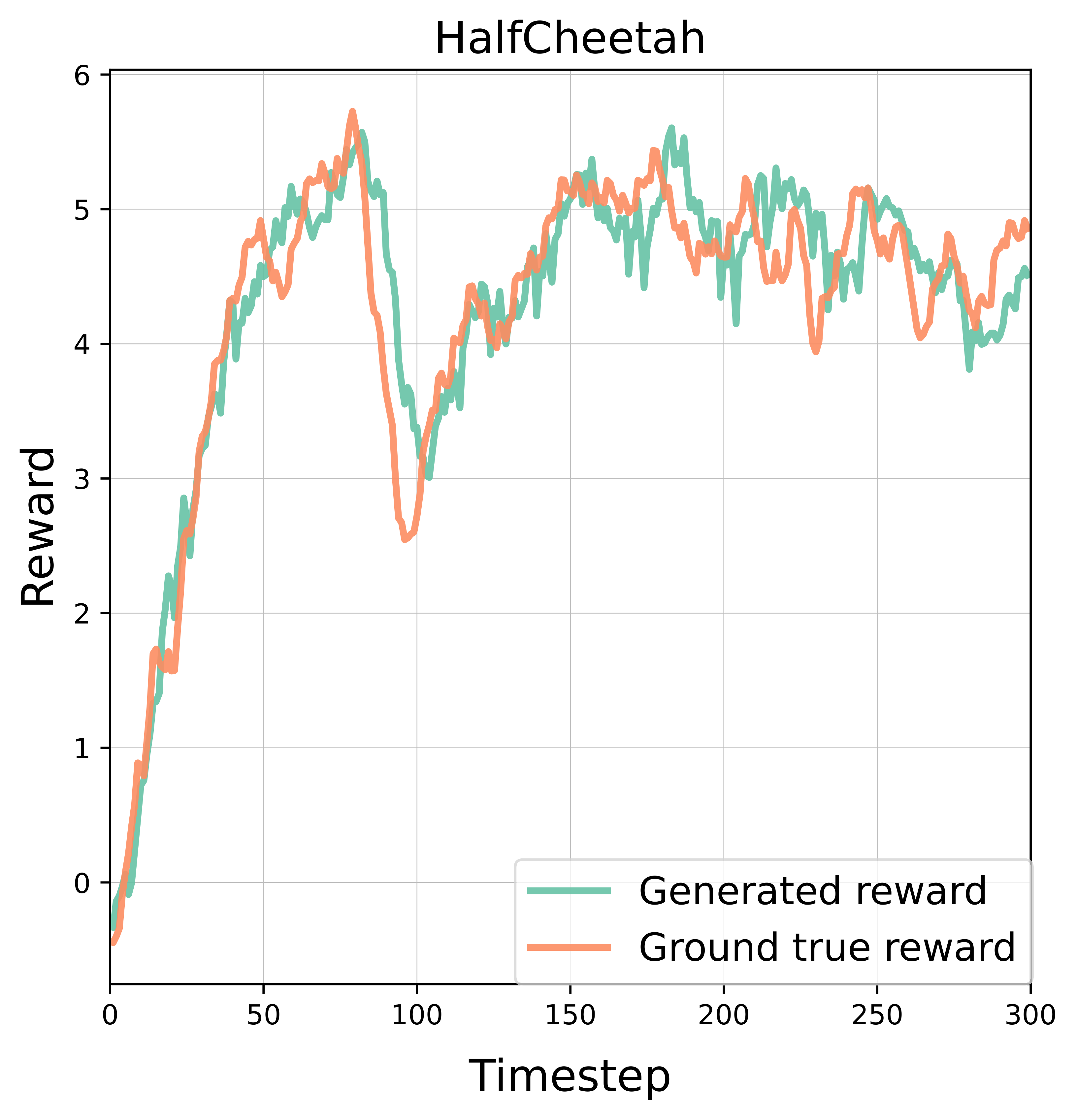}
    \includegraphics[height=4.2cm]{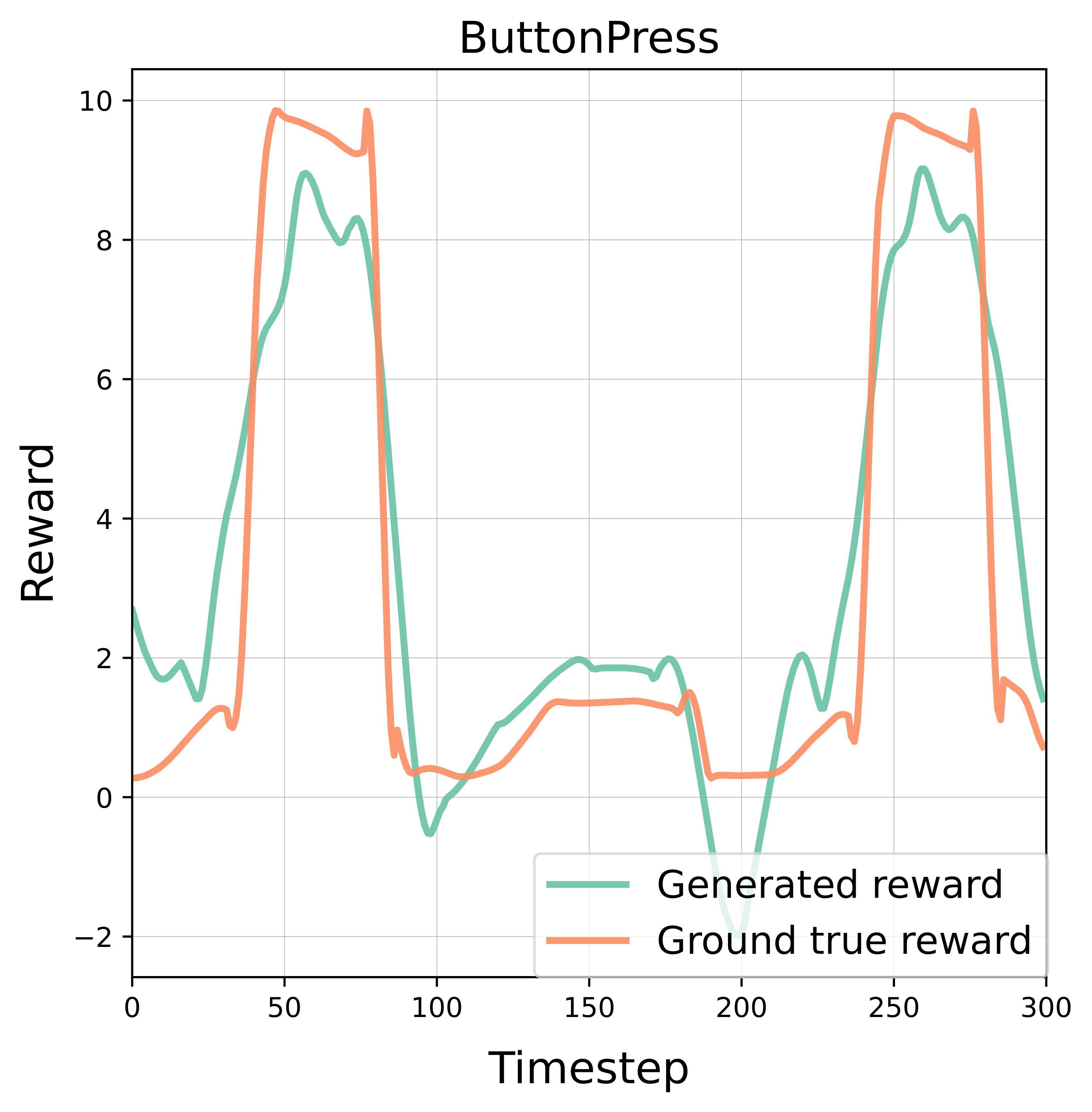}\label{4b}
    }
    \caption{(a) The episodic returns of learned reward and ground true reward (b) The learned reward signals and the ground true reward signals within a single episode along the timesteps.}
    \label{figure4} 
    \vspace{-16pt}
\end{figure}

\subsection{Experiment Setups}
\label{subsec:setups}
We compare our approach to previous methods to verify if our approach can achieve similar performance with less feedback. In Sec.~\ref{subsec: method comparisons}, we conduct ablation studies to investigate the influence of adaptive reward updates on the robustness of scoring errors, and to assess how various sampling methods affect performance. In Sec.~\ref{subsec:reward}, we analyze the learned reward and the agent's behavioral pattern to determine if our approach can accurately extrapolate the user's preferences and underlying intent. Finally, we conduct real human experiments in Sec.~\ref{subsec:real human}. 

We evaluate our proposed method on several continuous robotic tasks in simulation, including locomotion tasks Ant and HalfCheetah in Mujoco simulator \cite{todorov2012mujoco} with OpenAI Gym \cite{1606.01540}, and robotic manipulation tasks in Metaworld environment \cite{yu2019meta}, namely PushButton and SweepInto. For locomotion tasks, we use the episode return as the evaluation metric. For manipulation tasks, we use the task success rate of the last 100 episodes as the evaluation metric. We train $2,000$ episodes for Mujoco locomotion tasks and $3,000$ episodes for Metaworld robotic manipulation tasks each run. The episode is $300$ steps long, with the exception of the SweepInto task, which is $250$ steps long to reduce the proportion of task-goal-unrelated steps in the episode. 

We use the state-of-the-art off-policy DRL algorithm SAC to learn the behavioral policy \cite{haarnoja2018soft}. However, the agent can only receive the reward generated by our learned reward function. To model our reward function, we use a single deep neural network consisting of 3 fully connected layers of 256 units with leaky ReLUs. We train the reward network from scores using the Adam optimizer with a learning rate of $10^{-3}$ and a batch size of $128$.
We use a standard scoring range of 0 to 10 across all experiments. To evaluate our approach quantitatively, we make the agent learn tasks only by intermittently getting its past experience scored by a scripted teacher. By linearly mapping the episode returns to the scoring range, the scripted teacher provides scores. We use a two-stage scoring frequency: at the beginning of training, we use a faster scoring frequency, scoring 5 trajectories every 10 episodes. When the agent's performance reaches approximately a quarter of the maximum episode returns, we switch to a slower scoring frequency, scoring 10 trajectories every 100 episodes. For all experiments, we set the $\beta$ for adaptive sampling to $3$.

\subsection{Results}
\label{subsec: method comparisons}
To examine the effectiveness of our approach, we compare it to the original SAC algorithm training with the same ground true reward as we used for the scripted teacher. We use the same hyperparameters for both trainings. We also compare to the state-of-art preference learning algorithm PEBBLE \cite{lee2021pebble}. We use the exact same values of hyperparameters for PEBBLE as the Equal SimTeacher setting reported in \cite{lee2021b} and the corresponding open-source code repository.

Fig.~\ref{figure2} shows the learning curves of our approach and PEBBLE with different numbers of teacher preference feedback in comparison to SAC with true reward. Note that our approach employs a different type of feedback than PEBBLE. In this experiment setting, PEBBLE assigns a preference label to a pair of 50-step partial trajectories for single teacher feedback, whereas our approach assigns a global score to an entire episode with 300 steps. As a result, we give PEBBLE an advantage by providing more than three times as much feedback as ours. We can see that our approach achieves the same or higher level of performance than PEBBLE, which is given more feedback, in all tasks. In comparison to the SAC with ground true reward, our approach requires more training time to converge because it must learn the reward from scratch at the beginning of training, but it can match the performance after convergence in all tasks using only a small number of trajectory scores from the teacher. The results show that our approach can learn robot behavioral policies effectively in sparse reward environments with teachers' scores.

\subsection{Ablation Study}
\label{subsec: ablation study}
\subsubsection{Robustness to Scoring Errors}

\begin{figure}[!t]
\centering
\includegraphics[width=0.8\linewidth]{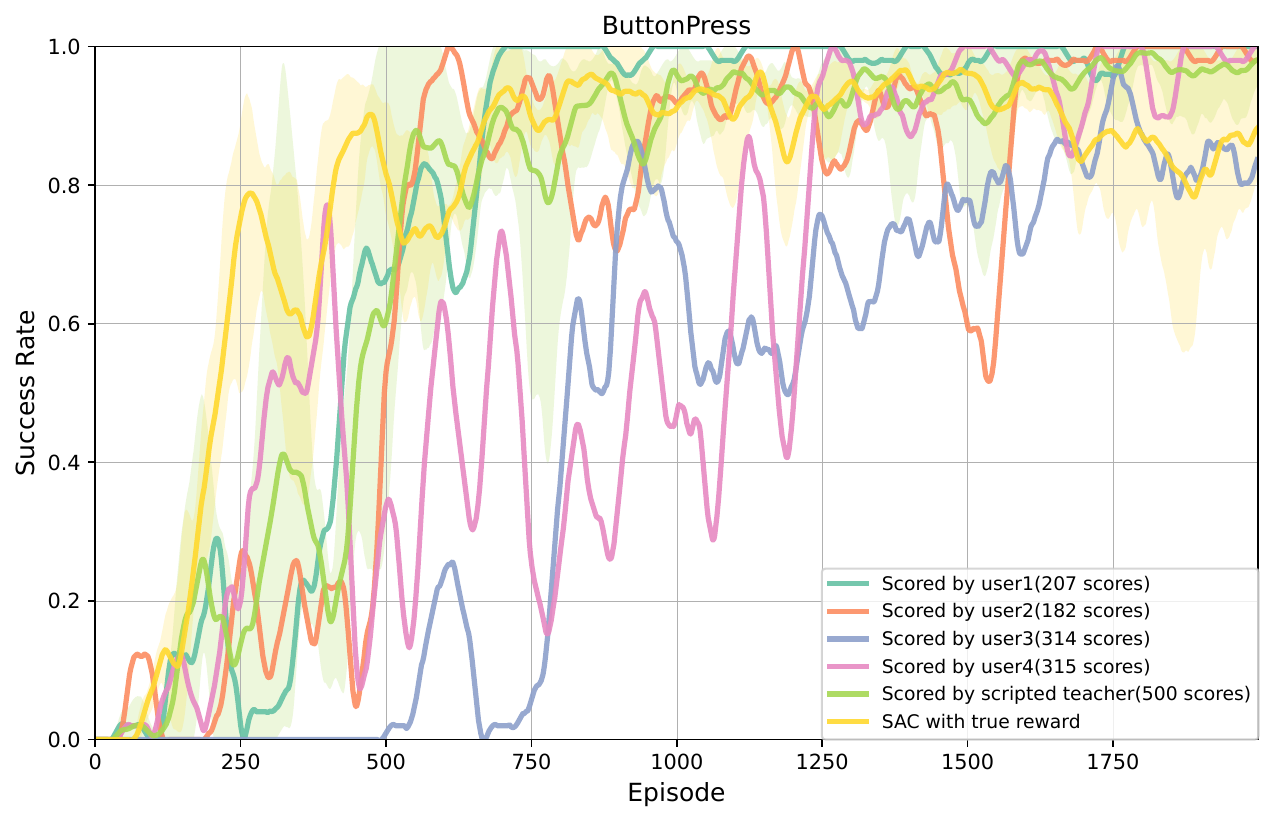}
\caption{The performance comparisons of original SAC and our approach trained by real human users and scripted teacher.}
\label{figure5}
\vspace{-20pt}
\end{figure}

The preceding experiments assume access to the perfect correct scores generated by the ground true reward. However, in practice, it is impossible for a human teacher to score hundreds of trajectories accurately: users can give vague and approximate scores for trajectories with similar performance. Thus, we examine the robustness to scoring errors and low scoring precision of our approach by comparing the performance of the noisy scores when using hard reward update, soft reward update by label smoothing, and adaptive reward update.


We simulate the real human teacher by adding noise randomly generated by Gaussian distribution to the scores given by the scripted teacher as $s^\prime = \mathcal{N}(s, \sigma_{noise}^{2})$. We adopt a minimal step of 0.5 for these noise-infused scores, such as 3.0 and 7.5. This permits teachers to give scores to trajectories that perform similarly. We tested our method with $\sigma_{noise}^{2} = 0.4$ and $\sigma_{noise}^{2} = 0.8$, and use Kendall's $\tau_{B}$ coefficient to further measure the rank correlation between the noisy scores and the perfect correct scores. This coefficient is calculated by $\tau_{B}=(P-Q)/\sqrt{(P + Q + T)(P + Q + U)}$, where $P$ is the number of concordant pairs, $Q$ is the number of discordant pairs, $T$ is the number of ties only in one group of data, and $U$ is the number of ties only in another, $\tau_{B}$ will be high and close to 1 if two variables have similar rank. We found that in our experiment, $\sigma_{noise}^{2} = 0.4$ corresponds to $\tau_{B} \approx 0.8$. The higher noise $\sigma_{noise}^{2} = 0.8$ leads to a lower correlation level $\tau_{B} \approx 0.65$. 

The results of training with imperfect scores on the HalfCheetah and ButtonPress tasks are shown in Fig.~\ref{3a}, where the smoothing strength is $\alpha = 0.05$ for the original label smoothing, $\alpha^\prime = 2$ for the adaptive reward update. Despite a slight decrease in performance compared to perfect scoring, the adaptive update method performs better compared to the other two methods in both tasks when scoring errors $\sigma_{noise}^{2} = 0.4$. With $\sigma_{noise}^{2} = 0.8$, the adaptive reward update method surpasses others in the ButtonPress task. However, in the relatively simple task, HalfCheetah did not gain extra advantages over the hard reward update. Overall, our proposed adaptive reward update method delivers the strongest performance and shows strong robustness to high-scoring errors.
We also investigate the effects of different sampling methods to select scored trajectory pairs for reward updates. We used $\sigma_{noise}^{2} = 0.4$ to simulate the real scoring scenario. Fig.~\ref{3b} shows the learning curves of our approach on the HalfCheetah and ButtonPress tasks under three different sampling schemes: uniform sampling, entropy-based sampling, and priority-based sampling. We can see that the priority-based sampling method significantly outperforms other sampling methods. Although entropy-based sampling performs well with a perfect feedback teacher as suggested in \cite{lee2021b}, it cannot handle the noisy-scoring scenario well.

\subsection{Reward Extrapolation} 
\label{subsec:reward}
\subsubsection{Reward Analysis}
We compare the learned reward function to the ground truth rewards to assess the quality of the learned reward function. We run SAC with true reward to collect trajectories with a variety of performance qualities, and then we compare the episodic ground truth returns to the returns generated by the learned reward function. Fig.~\ref{4a} shows the reward function learned by our approach on 250 scores and 500 scores in HalfCheetah and ButtonPress respectively. We can see that the learned reward function has a strong correlation with the true reward on the ground. It should be noted that the learned and true rewards have very different scales, but this difference had no effect on policy learning performance. We further investigate the rewards by looking into the reward functions within an episode at different timesteps. We generate a set of suboptimal trajectories with high and low reward ranges in one episode. The results are shown in Fig.~\ref{4b}. We manually normalize the learned reward outputs to have the same scale as the true rewards by multiplying a coefficient. We can see that the learned rewards are well-aligned with the ground truth rewards. 

\subsubsection{Customized Behavior}
One goal of our approach is to enable users to train customized policies through scoring. We demonstrate this in the RLBench \cite{james2019rlbench} simulation task PushButton, which requires a Franka Emika Panda robot arm to push a button on a table. We model two scripted teachers to score trajectories with different preferences as follows: (1) teacher 1: robot first moves to the top of the button, then pushes with the gripper tip while remaining vertical, (2) teacher 2: robot moves its gripper parallel to the table and presses the button with its side. For the trained agent's policies please refer to the supplementary video. The result demonstrates that our method can infer users' underlying intent and complete tasks in accordance with the user's preferences.

\subsection{Real Human Experiment}
\label{subsec:real human}
We conduct experiments with real human users to test our approach. We create a graphical user interface (GUI) and test it with two users on the MetaWorld ButtonPress environment. The GUI displays four previously scored trajectories and their scores as references to help users score new trajectories consistently. We select two scored trajectories with the closest predicted returns to the current trajectory and two references that are most similar to the current trajectory in Cartesian space, measured by dynamic time warp (DTW) \cite{berndt1994using}. Users are allowed to revise the scores of the reference trajectories as needed, and they could skip scoring if they find it difficult. We follow the two-stage scoring frequency outlined in Sec.~\ref{subsec:setups}, starting with a faster scoring frequency and allowing users to switch to a lower frequency mode based on their performance. Fig.~\ref{figure5} shows the learning curve of the four users compared to learning by SAC with true reward and learning by our approach with a scripted teacher. Our approach shows that a good behavior policy could be trained with only about three hundred scores. For more information on using the scoring interface, please refer to the supplementary video.



\section{Conclusion}

We propose an algorithm for interactive RL that uses scores from a teacher to learn both a policy and reward function. This eliminates the need for human demonstrations and maximizes the use of user feedback, reducing the amount of required feedback. Our experiments show that even with a small number of human scores, our method can train robotic locomotion and manipulation tasks to near-optimal levels. With this method, we can map global behavior evaluations to rewards for only states or state-action pairs, allowing us to learn optimal policies in environments where rewards cannot be observed.


\footnotesize
\bibliographystyle{IEEEtran} 
\bibliography{oprrl}

\end{document}